\documentclass[11pt]{article}
\usepackage{coling2020}
\usepackage{times}
\usepackage{url}
\usepackage[utf8]{inputenc}
\usepackage{latexsym}
\usepackage{graphicx}
\usepackage{xcolor}
\usepackage{hyperref}
\usepackage{comment}

\newcommand{\MYhref}[3][blue]{\href{#2}{\color{#1}{#3}}}%

 \colingfinalcopy 

\title{\texttt{comp-syn}: Perceptually Grounded Word Embeddings with Color}

\author{Bhargav Srinivasa Desikan \\
  University of Chicago \\
  Knowledge Lab  \\
  {\tt bhargav@uchicago.edu} \\\And
     
  Tasker Hull  \\
  Psiphon Inc, Toronto \\
  {\tt t.mackersy@psiphon.ca} \\\AND
  
  Ethan O. Nadler \\
  Stanford University \\
  KIPAC \& Department of Physics \\
  {\tt enadler@stanford.edu} \\   \And
  
  Douglas Guilbeault  \\
  University of California, Berkeley \\
  Haas Business School  \\

  {\tt douglas.guilbeault@berkeley.edu} \\\AND

   Aabir Abubaker Kar  \\
  University of Chicago \\
  Knowledge Lab  \\
  {\tt aabir@uchicago.edu} \\\And
   
       Mark Chu  \\
  Columbia University \\
  {\tt mbc2165@columbia.edu} \\ \And

  Donald Ruggiero Lo Sardo  \\
  Sony CSL Paris \\
  {\tt losardor@gmail.com} \\
  } 

\date{}

\begin{document}
\maketitle
\vspace{-12mm}
\begin{abstract}

Popular approaches to natural language processing create word embeddings based on textual co-occurrence patterns, but often ignore embodied, sensory aspects of language. Here, we introduce the Python package \texttt{comp-syn}, which provides grounded word embeddings based on the perceptually uniform color distributions of Google Image search results. We demonstrate that \texttt{comp-syn} significantly enriches models of distributional semantics. In particular, we show that (1) \texttt{comp-syn} predicts human judgments of word concreteness with greater accuracy and in a more interpretable fashion than \texttt{word2vec} using low-dimensional word--color embeddings, and~(2) \texttt{comp-syn} performs comparably to \texttt{word2vec} on a metaphorical vs.\ literal word-pair classification task. \texttt{comp-syn} is open-source on PyPi and is compatible with mainstream machine-learning Python packages. Our package release includes word--color embeddings for over 40,000 English words, each associated with crowd-sourced word concreteness judgments. 

\end{abstract}

\vspace{1mm}


\section{Introduction}

The embodied cognition paradigm 
seeks to ground semantic processing in bodily, affective, and social experiences. This paradigm explains how sensory information contributes to semantic processing, either through metaphors involving references to sensory experience \cite{lakoff1989more,gallese2005brain} or through the simulation of sensory experience in mental imagery \cite{bergen2012louder}. For example, color is pervasive in linguistic metaphors (e.g., ``her bank accounts are in the \emph{red}''), and primes a range of affective and interpretative responses \cite{Mehta2009,Elliot2014}. 

Extant methods in computational linguistics are limited in their ability to advance the embodied cognition paradigm due to their focus on text, which often precludes multi-modal analyses of images and other forms of sensory data---including color---involved in human meaning-making activities. \textit{Distributional semantics} is a particularly prominent approach, wherein textual co-occurrence patterns are used to embed words in a high-dimensional space; the resulting ``distance'' between word embeddings correlates with semantic similarity \cite{landauer1997solution}. Although leveraging neural networks~\cite{mikolov2013distributed} and syntactic information \cite{levy2014dependency} has led to recent progress, popular models like \texttt{word2vec} lack firm grounding in their use of sensory information. Moreover, high-dimensional word embeddings created by neural networks are difficult to interpret \cite{senel_interp} and require long training periods \cite{Ji_training}. Thus, it remains difficult to reconcile models of distributional semantics with theories of embodied semantics.

To address these limitations, \emph{multi-modal} approaches to distributional semantics combine text and image data. However, these models continue to infer semantic associations using high-dimensional word-plus-image embeddings; their interpretability therefore remains an issue \cite{bruni2014multimodal,socher2014grounded,lu2019vilbert}. With respect to color, multi-modal models operate in standard colorspaces like RGB that are not perceptually uniform \cite{CIE} and embed color in a manner that combines it with complex, multidimensional spatial information.

Here, we introduce a novel word embedding method based on color that is explicitly interpretable with respect to theories of embodied cognition. We build on work which shows that color distributions in online images reflect both affective and semantic similarities among words in abstract domains (e.g., in the domain of academic disciplines), while also characterizing human judgments of concept concreteness \cite{brysbaert2014concreteness,guilbeault2020color}. We present the Python package \texttt{comp-syn}\footnote{Short for ``Computational Synaesthesia.''}, which allows users to explore word--color embeddings based on the perceptually uniform color distributions of Google Image search results. We provide embeddings for a set of 40,000 common English words, and we benchmark the performance of our model using crowd-sourced human concreteness ratings \cite{brysbaert2014concreteness}. We show that \texttt{comp-syn} complements the performance of text-based distributional semantics models by providing an interpretable embedding that both (1) predicts human judgments of concept concreteness, and (2) distinguishes metaphorical and literal word pairs.

\section{Python Package: \texttt{comp-syn}}

\MYhref{https://github.com/comp-syn/comp-syn}{\texttt{comp-syn}} is an open-source Python package available on GitHub and downloadable through PyPi. The code follows Python best practices and uses industry standard packages for scientific computing, facilitating easy integration with Python code bases; we provide complete details in the Supplementary Material. \texttt{comp-syn} creates word--color embeddings by computing the perceptually uniform $J_z A_z B_z$ \cite{safdar2017perceptually} color distributions of their corresponding top $100$ Google Image search results. We represent these distributions by their mean and (optionally) standard deviation in $8$ evenly-segmented $J_z A_z B_z$ bins, yielding 8 to 16-dimensional word--color embeddings. The details of our code, image searches and methods for generating word--color embeddings are described in the Appendices.

We use Google rather than a curated image database such as ImageNet \cite{Deng2009} because popular datasets are heavily biased toward concrete objects, which limits their applicability in abstract semantic domains. Moreover, Google Image search results reflect content that users interact with most frequently \cite{jing2008pagerank}, underscoring their relevance for connecting distributional properties of words and images to human semantic processing. To ensure that our approach is entirely unsupervised, we do not select particular features of images when measuring their color distributions. This approach avoids importing pre-determined semantic notions into our analysis and connects our method to cognitive theories that attribute aesthetic relevance to recognizable features in both the background and foreground of images \cite{riley1995,Elliot2014,guilbeault2020color}.

\begin{table*}[t]
\begin{center}
\begin{tabular*}{\textwidth}{l l l l l}
Model & Embedding based on & Dimension & Concreteness prediction & Metaphor task \\
\hline
\hline\\[-0.5em]
\texttt{comp-syn} & \vtop{\hbox{\strut Perceptually uniform}\hbox{\strut color distributions~~}} & $8$--$16$ & \vtop{\hbox{\strut Linear: $R^2=0.96$}\hbox{\strut Nonlinear: $\ln \mathcal{L}=86$~~}} & \vtop{\hbox{\strut $92\%$ test}\hbox{\strut set accuracy~~}} \vspace{1mm} \\
\texttt{word2vec} & Textual co-occurrence & $\sim 300$ & \vtop{\hbox{\strut Linear: $R^2=0.17$}\hbox{\strut Nonlinear: $\ln\mathcal{L}=76$~~}} & \vtop{\hbox{\strut $95\%$ test}\hbox{\strut set accuracy~~}} \\
\hline
\end{tabular*}
\end{center}
\caption{\label{tab:table}
Comparison of \texttt{comp-syn} and \texttt{word2vec} and summary of our main results.
}
\end{table*}

\section{Results}

Table \ref{tab:table} summarizes our main results, which we now describe in turn.

\begin{figure}[t]
\centering
    \includegraphics[width=0.825\textwidth]{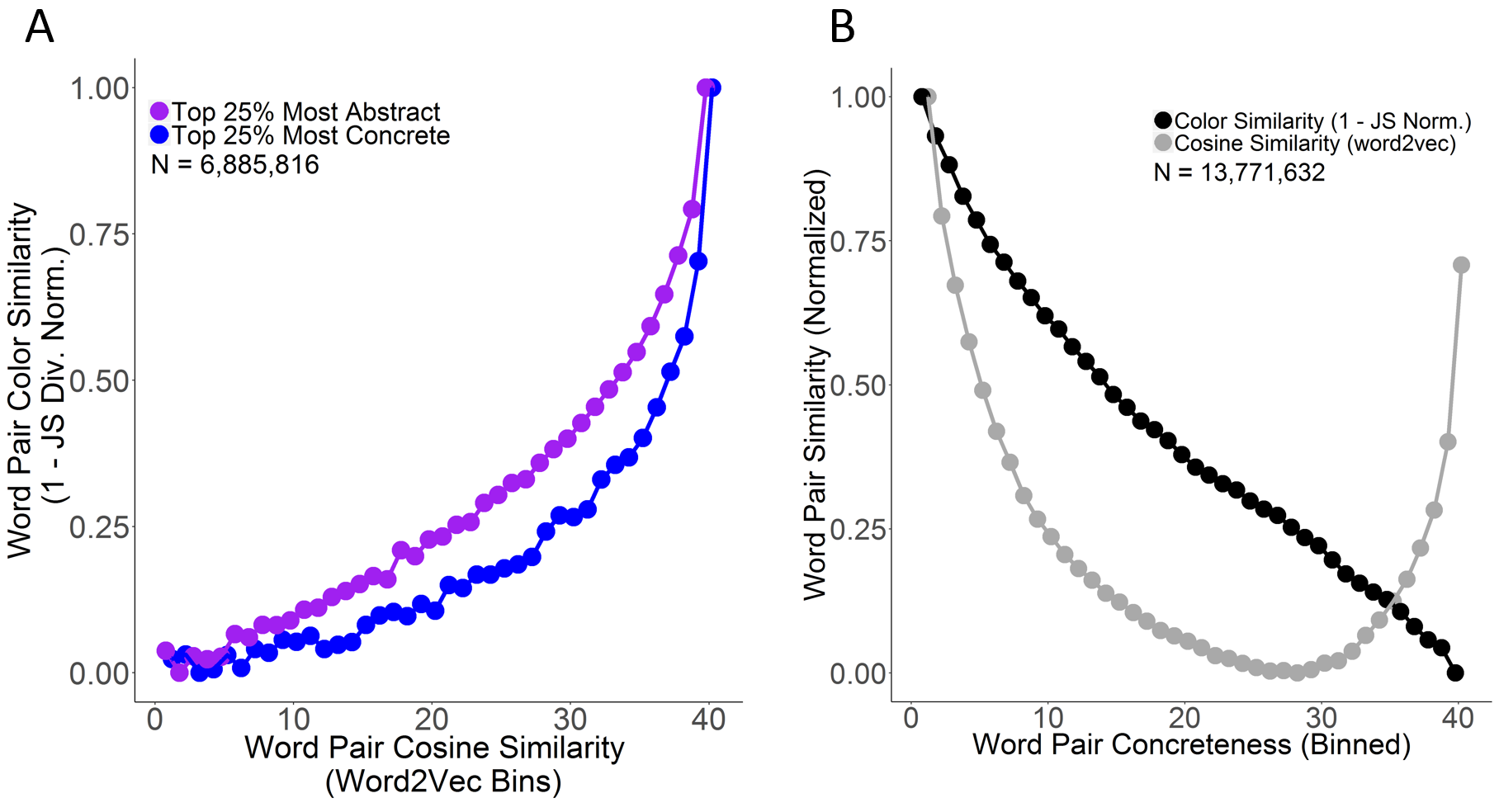}
    \caption{The semantic coherence of \texttt{comp-syn}. (A) Word pair color similarity in \texttt{comp-syn}, measured using the Jensen-Shannon divergence between their perceptually uniform color distributions, correlates with \texttt{word2vec} cosine similarity ($N>10^6$ pairs). This holds for both concrete (blue) and abstract (magenta) word pairs, rated according to crowd-sourced human judgments. Data represent average color similarity in $40$ equally-sized bins of \texttt{word2vec} similarity. (B) Word pair similarity vs.\ human concreteness judgments ($N >10^7$ pairs). \texttt{comp-syn} similarity monotonically decreases with word pair concreteness, while \texttt{word2vec} similarity is a nonlinear function of word pair concreteness. Data represent average word pair similarity in \texttt{word2vec} (gray) and \texttt{comp-syn} (black) in $40$ equally-sized bins of summed word pair concreteness.}
    \label{fig:fig1}
\end{figure}

\subsection{Comparison to \texttt{word2vec} Similarity}

We begin by demonstrating that pairwise word similarity is highly correlated in \texttt{comp-syn} and \texttt{word2vec}, illustrating the semantic coherence of our color embeddings. We calculate pairwise distances between each of the 40,000 words in our dataset and 500 randomly-selected words from the same set, yielding over $10^7$ distinct pairs. We compare the cosine similarity between word pairs in the Mikolov et al.\ \shortcite{mikolov2013distributed} \texttt{word2vec} model with the Jensen-Shannon (JS) divergence between $J_z A_z B_z$ distributions in \texttt{comp-syn}. Fig.\ 1A shows that JS divergence in \texttt{comp-syn} is significantly correlated with cosine similarity in \texttt{word2vec} ($p < 0.00001$, $\mathrm{JT}=773$, Jonckheere-Terpstra test). This implies that our low-dimensional, interpretable embedding captures aspects of the key information contained in a widely-used distributional semantics model.

\subsection{Relation to Human Concreteness Judgments}

Next, we examine the relation between our word--color embeddings and human concreteness judgments \cite{brysbaert2014concreteness}. We label word pairs with summed concreteness ratings in the highest (lowest) quartile of our data as ``concrete'' (``abstract''). Fig.\ 1A shows that abstract word pairs are more similar in colorspace than concrete word pairs, even at fixed \texttt{word2vec} similarity 
($p = 0.001$, $\mathrm{DF} = 77$, Dickey-Fuller test). Moreover, as shown in Fig.\ 1B, \texttt{comp-syn} captures concreteness judgments in a more interpretable fashion than \texttt{word2vec}. In particular, because abstract words are more similar in colorspace (on average), there is a monotonic relationship between color similarity and concreteness; indeed, a linear model of \texttt{comp-syn} similarity accounts for nearly all of the variance in word pair concreteness ($R^2=0.96$). On the other hand, \texttt{word2vec} similarity is a nonlinear function of word pair concreteness. Although nonlinear functions of \texttt{word2vec} similarity also predict concreteness well~($R^2>0.99$), \texttt{comp-syn} versions of the same nonlinear models are significantly more accurate (log-likelihood difference $\Delta \ln \mathcal{L}\sim 10$ in favor of \texttt{comp-syn}) and less complex (Bayesian information criterion $\Delta\mathrm{BIC}\sim 20$, also in favor of \texttt{comp-syn}).

To qualitatively explore the relationship between our word--color embeddings and human concreteness judgment, Fig.\ 3A shows the perceptually uniform color distributions associated with some of the most and least concrete words in our corpus. These \emph{colorgrams} illustrate that concrete words (e.g., ``pyramid'') are often associated with color distributions that are peaked in specific regions of colorspace, while abstract words (e.g., ``concept'') feature more variegated color distributions. The spatial and textural features of the images reflect these properties, and exploring the relationship between these aspects of \emph{colorgrams}, is an interesting avenue for future study.



\begin{figure}[t]
\hspace{-2.5mm}
    \includegraphics[trim=0 2cm 0 0,width=1.03\textwidth]{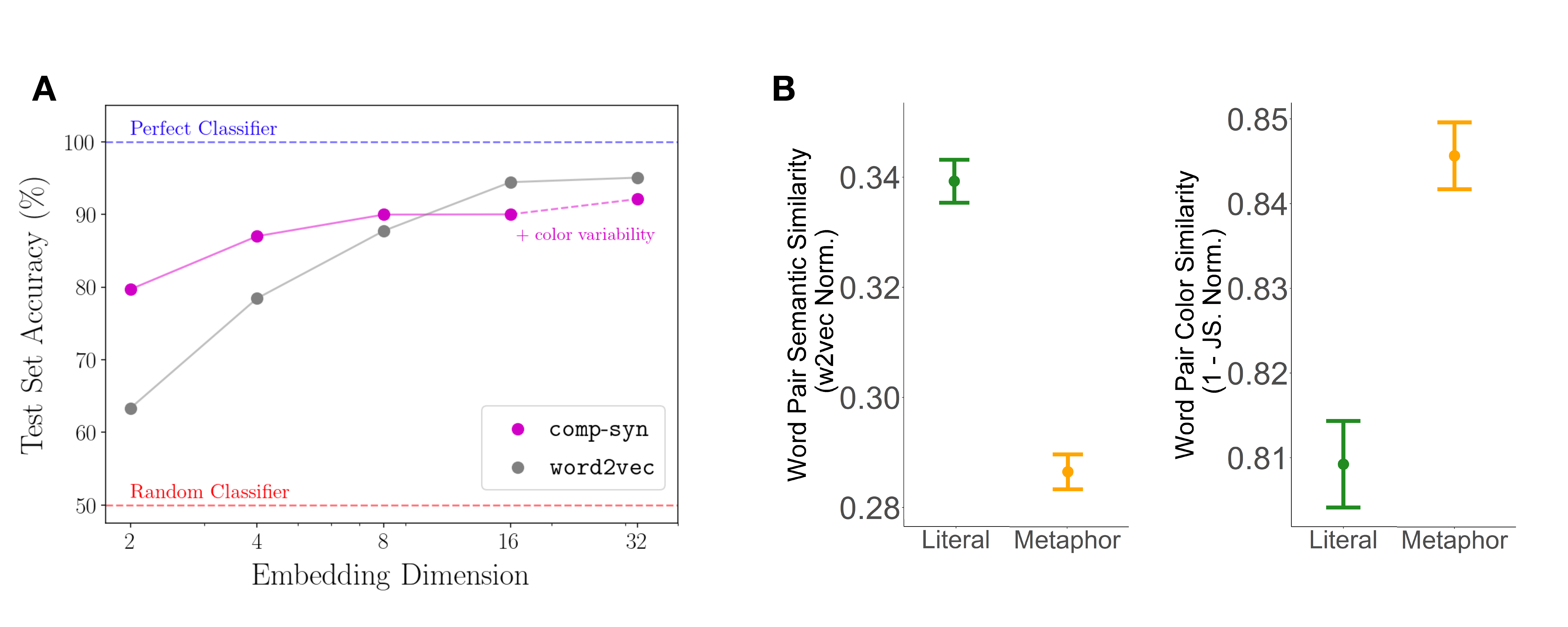}
    \caption{Classifying metaphorical vs.\ literal adjective-noun pairs with \texttt{comp-syn}. (A) Test set classification accuracy for \texttt{comp-syn} (magenta) and \texttt{word2vec} (gray) as a function of PCA embedding dimension. (B) Average similarity of metaphorical and literal adjective-noun pairs in \texttt{word2vec} (left panel) and \texttt{comp-syn} (right panel). Error bars indicate $95\%$ confidence intervals.}
    \label{fig:fig2}
\end{figure}

\subsection{Metaphor Pair Analysis}

The results above suggest that our word--color embeddings encode complementary information about concept concreteness relative to purely textual embeddings. This raises the question of what can be learned from cases in which \texttt{word2vec} and \texttt{comp-syn} provide conflicting similarity predictions when evaluated on the same pair of words. Here, we address this question by demonstrating that \texttt{comp-syn} significantly enriches metaphorical word pair classification, which often requires extensive manual tagging due to subtle uses of both sensory and abstract features \cite{lakoff2008metaphors,Bethard2009,Bipin2013,Dodge2015,Winter2019}.


We trained a gradient-boosted tree classifier implemented via \texttt{XGBoost} \cite{Chen:2016} to label adjective-noun pairs as either metaphorical or literal, using over $8000$ word pairs from Tsvetkov et al.\ \shortcite{tsvetkov-etal-2014-metaphor} and Guti{\'e}rrez et al.\ \shortcite{gutierrez-etal-2016-literal}. This dataset encompasses a statistically representative range of metaphorical and literal contexts for each adjective \cite{gutierrez-etal-2016-literal}. To compare embeddings, we compressed \texttt{word2vec}'s 300-dimensional word vector differences using PCA to match the dimensionality of \texttt{comp-syn}, following Bolukbasi et al.\ \shortcite{Bolukbasi_nips}. Fig.\ 2A shows that a classifier trained using only \texttt{word2vec} achieves a limiting test set accuracy of $95\%$, compared to $92\%$ for \texttt{comp-syn}. Importantly, \texttt{comp-syn} outperforms \texttt{word2vec} at low embedding dimensions, indicating that it captures semantic content in an interpretable fashion. This analysis does not demonstrate either model's best-case performance on this task; rather, it highlights the complementary information provided by \texttt{comp-syn}.

Strikingly, \texttt{word2vec} and \texttt{comp-syn} distances provide qualitatively different information when distinguishing literal and metaphorical adjective-noun pairs. Fig.\ 2B shows that literal adjective-noun pairs are more similar than metaphorical pairs in \texttt{word2vec} ($p<0.001$, Wilcoxon rank sum); the reverse holds in \texttt{comp-syn}, where literal pairs are significantly \emph{less} similar than metaphorical pairs ($p<0.001$, Wilcoxon rank sum). This is a consequence of the fact that images associated with concrete words are more variable in colorspace \cite{guilbeault2020color}. In this way, \texttt{comp-syn} reveals differences in color similarity between literal and metaphorical adjective-noun pairs that are of interest for cognitive theory \cite{Bipin2013}. Particularly, our findings suggest that metaphors can exploit color similarities between words that are dissimilar in textual embeddings, which may help facilitate cognitive processing of semantic relations among concepts from distinct domains \cite{guilbeault2020color}.

Qualitative inspection of specific adjective-noun word pairs highlights some notable differences between textual and word--color embeddings. Fig.\ 3B provides visual representations of the color distributions for word pairs in our metaphorical vs.\ literal dataset that are most and least similar in \texttt{comp-syn}. Interestingly, while metaphorical pairs are more similar than literal pairs in \texttt{comp-syn}, the \emph{least} similar metaphorical pairs explicitly invoke color, e.g., ``deep orange''. Algorithmically, this is due to the fact that \texttt{comp-syn} embeddings associated with color terms are unusually coherent. On the other hand, the color distributions associated with the most similar pairs in \texttt{word2vec} (e.g., ``bushy beard'') often noticeably contrast, while \texttt{word2vec}'s least similar pairs (e.g., ``rough customer'') do not strongly invoke color. These findings point to an important direction for future research enabled by the grounded nature of \texttt{comp-syn}: how do linguistic metaphors leverage sensory information to characterize colorspace itself (e.g., in the use of spatial information in the popular metaphor ``deep purple'')?

\begin{figure}[t]
\hspace{-14mm}
    \includegraphics[width=1.15\textwidth]{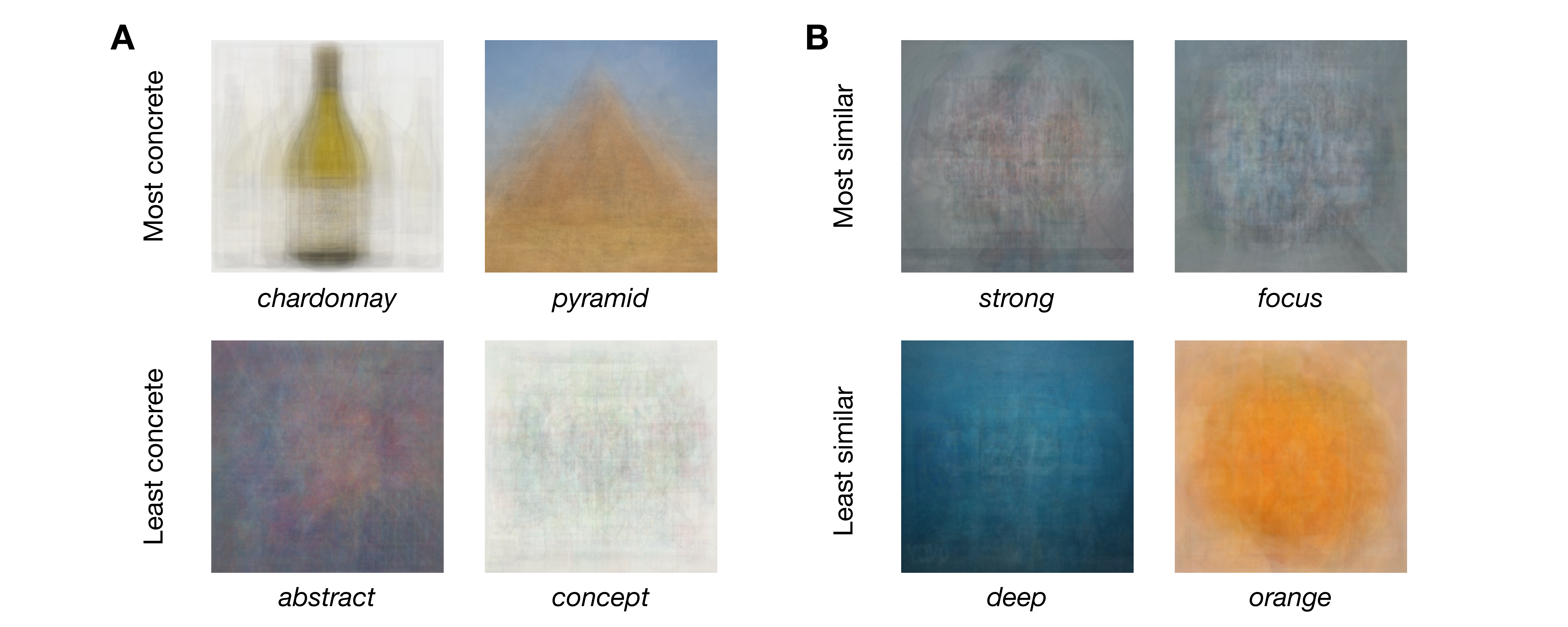}
    \caption{(A) Examples of the most and least concrete terms, visualized using our word--color embedding method. (B) Examples of the most and least similar adjective-noun pairs according to \texttt{comp-syn} word--color embeddings. We represent each term using a \emph{colorgram}, i.e., a composite image produced by averaging the perceptually uniform colors of pixels across Google Image search results.}
    \label{fig:fig3}
\end{figure}


\section{Conclusion and Future Work}
We have presented \texttt{comp-syn}, a new Python package that provides perceptually grounded color-based word embeddings. These embeddings are interpretable with respect to theories of embodied cognition because (1) \texttt{comp-syn} represents color in a fashion that emulates human perception, and~(2) \texttt{comp-syn} leverages Google Image search results that human users dynamically interact with and produce. By linking \texttt{comp-syn} with human concreteness judgments for 40,000 common English words, our package provides a multi-modal playground for exploring grounded semantics. We demonstrated that \texttt{comp-syn} enriches popular distributional semantics models in both word concreteness prediction and metaphorical word-pair classification. A myriad of \texttt{comp-syn} applications await, including color-based classification of text genres and the characterization of sensory imagery in everyday language.

\section*{Acknowledgements}

The authors gratefully acknowledge the support of the Complex Systems Summer School hosted at the Institute of American Indian Arts and the Santa Fe Institute, where this project was initiated. D.G. gratefully acknowledges financial support from the Institute on Research and Innovation in Science. This research received support from the National Science Foundation (NSF) under grant no. NSF DGE-1656518 through the NSF Graduate Research Fellowship received by E.O.N. D.G. and T.H. acknowledge intellectual support from the Institute for Advanced Learning (IAL) in Ontario, Canada.



\bibliography{coling2020.bib}

\begin{thebibliography}{}

\bibitem[\protect\citename{Bergen}2012]{bergen2012louder}
Benjamin~K Bergen.
\newblock 2012.
\newblock {\em Louder than words: The new science of how the mind makes
  meaning}.
\newblock Basic Books.

\bibitem[\protect\citename{Bethard \bgroup et al.\egroup }2009]{Bethard2009}
Steven Bethard, Vicky Lai, and James Martin.
\newblock 2009.
\newblock Topic model analysis of metaphor frequency for psycholinguistic
  stimuli.
\newblock {\em Proceedings of the Workshop on Computational Approaches to
  Linguistic Creativity}, pages 9--16.

\bibitem[\protect\citename{Bolukbasi \bgroup et al.\egroup
  }2016]{Bolukbasi_nips}
Tolga Bolukbasi, Kai-Wei Chang, James Zou, Venkatesh Saligrama, and Adam Kalai.
\newblock 2016.
\newblock Man is to computer programmer as woman is to homemaker? debiasing
  word embeddings.
\newblock In {\em Proceedings of the 30th International Conference on Neural
  Information Processing Systems}, NIPS’16, page 4356–4364, Red Hook, NY,
  USA. Curran Associates Inc.

\bibitem[\protect\citename{Bruni \bgroup et al.\egroup
  }2014]{bruni2014multimodal}
Elia Bruni, Nam-Khanh Tran, and Marco Baroni.
\newblock 2014.
\newblock Multimodal distributional semantics.
\newblock {\em Journal of artificial intelligence research}, 49:1--47.

\bibitem[\protect\citename{Brysbaert \bgroup et al.\egroup
  }2014]{brysbaert2014concreteness}
Marc Brysbaert, Amy~Beth Warriner, and Victor Kuperman.
\newblock 2014.
\newblock Concreteness ratings for 40 thousand generally known english word
  lemmas.
\newblock {\em Behavior research methods}, 46(3):904--911.

\bibitem[\protect\citename{Chen and Guestrin}2016]{Chen:2016}
Tianqi Chen and Carlos Guestrin.
\newblock 2016.
\newblock {XGBoost}: A scalable tree boosting system.
\newblock In {\em Proceedings of the 22nd ACM SIGKDD International Conference
  on Knowledge Discovery and Data Mining}, KDD '16, pages 785--794, New York,
  NY, USA. ACM.

\bibitem[\protect\citename{Clark}2018]{clark2018pillow}
Alex Clark.
\newblock 2018.
\newblock Pillow python imaging library.
\newblock {\em Pillow—Pillow (PIL Fork) 5.4. 1 documentation}.

\bibitem[\protect\citename{Deng \bgroup et al.\egroup }2009]{Deng2009}
Jia Deng, Wei Dong, Richard Socher, Li-Jia Li, Kai Li, and Li~Fei-Fei.
\newblock 2009.
\newblock Imagenet: A large-scale hierarchical image database.
\newblock {\em 2009 IEEE Conference on Computer Vision and Pattern
  Recognition}.

\bibitem[\protect\citename{Dodge \bgroup et al.\egroup }2015]{Dodge2015}
Ellen Dodge, Jisup Hong, and Elise Stickles.
\newblock 2015.
\newblock Metanet: Deep semantic automatic metaphor analysis.
\newblock {\em Proceedings of the Third Workshop on Metaphor in NLP}, pages
  40--49.

\bibitem[\protect\citename{Elliot and Maier}2014]{Elliot2014}
Andrew Elliot and Markus Maier.
\newblock 2014.
\newblock Color psychology: Effects of perceiving color on psychological
  functioning in humans.
\newblock {\em Annual Review of Psychology}, 65:95--120.

\bibitem[\protect\citename{Gallese and Lakoff}2005]{gallese2005brain}
Vittorio Gallese and George Lakoff.
\newblock 2005.
\newblock The brain's concepts: The role of the sensory-motor system in
  conceptual knowledge.
\newblock {\em Cognitive neuropsychology}, 22(3-4):455--479.

\bibitem[\protect\citename{Guilbeault \bgroup et al.\egroup
  }2020]{guilbeault2020color}
Douglas Guilbeault, Ethan~O Nadler, Mark Chu, Donald Ruggiero~Lo Sardo,
  Aabir~Abubaker Kar, and Bhargav~Srinivasa Desikan.
\newblock 2020.
\newblock Color associations in abstract semantic domains.
\newblock {\em Cognition}, 201:104306.

\bibitem[\protect\citename{Guti{\'e}rrez \bgroup et al.\egroup
  }2016]{gutierrez-etal-2016-literal}
E.~Dario Guti{\'e}rrez, Ekaterina Shutova, Tyler Marghetis, and Benjamin
  Bergen.
\newblock 2016.
\newblock Literal and metaphorical senses in compositional distributional
  semantic models.
\newblock In {\em Proceedings of the 54th Annual Meeting of the Association for
  Computational Linguistics (Volume 1: Long Papers)}, pages 183--193, Berlin,
  Germany, August. Association for Computational Linguistics.

\bibitem[\protect\citename{Hunter}2007]{hunter2007matplotlib}
John~D Hunter.
\newblock 2007.
\newblock Matplotlib: A 2d graphics environment.
\newblock {\em Computing in science \& engineering}, 9(3):90--95.

\bibitem[\protect\citename{Indurkhya and Ojha}2013]{Bipin2013}
Bipin Indurkhya and Amitash Ojha.
\newblock 2013.
\newblock An empirical study on the role of perceptual similarity in visual
  metaphors and creativity.
\newblock {\em Metaphor and Symbol}, 28.

\bibitem[\protect\citename{{International Commission On
  Illumination}}1978]{CIE}
{International Commission On Illumination}.
\newblock 1978.
\newblock Recommendations on uniform color spaces, color-difference equations,
  psychometric color terms.
\newblock {\em Color Research \& Application}, 15.

\bibitem[\protect\citename{{Ji} \bgroup et al.\egroup }2019]{Ji_training}
S.~{Ji}, N.~{Satish}, S.~{Li}, and P.~K. {Dubey}.
\newblock 2019.
\newblock Parallelizing word2vec in shared and distributed memory.
\newblock {\em IEEE Transactions on Parallel and Distributed Systems},
  30(9):2090--2100.

\bibitem[\protect\citename{Jing and Baluja}2008]{jing2008pagerank}
Yushi Jing and Shumeet Baluja.
\newblock 2008.
\newblock Pagerank for product image search.
\newblock In {\em Proceedings of the 17th international conference on World
  Wide Web}, pages 307--316.

\bibitem[\protect\citename{Jones \bgroup et al.\egroup }2001]{jones2001scipy}
Eric Jones, Travis Oliphant, Pearu Peterson, et~al.
\newblock 2001.
\newblock Scipy: Open source scientific tools for python.

\bibitem[\protect\citename{Lakoff and Johnson}2008]{lakoff2008metaphors}
George Lakoff and Mark Johnson.
\newblock 2008.
\newblock {\em Metaphors we live by}.
\newblock University of Chicago press.

\bibitem[\protect\citename{Lakoff and Turner}1989]{lakoff1989more}
George Lakoff and Mark Turner.
\newblock 1989.
\newblock {\em More than cool reason: A field guide to poetic metaphor}.
\newblock University of Chicago press.

\bibitem[\protect\citename{Landauer and Dumais}1997]{landauer1997solution}
Thomas~K Landauer and Susan~T Dumais.
\newblock 1997.
\newblock A solution to plato's problem: The latent semantic analysis theory of
  acquisition, induction, and representation of knowledge.
\newblock {\em Psychological review}, 104(2):211.

\bibitem[\protect\citename{Levy and Goldberg}2014]{levy2014dependency}
Omer Levy and Yoav Goldberg.
\newblock 2014.
\newblock Dependency-based word embeddings.
\newblock In {\em Proceedings of the 52nd Annual Meeting of the Association for
  Computational Linguistics (Volume 2: Short Papers)}, pages 302--308.

\bibitem[\protect\citename{Lu \bgroup et al.\egroup }2019]{lu2019vilbert}
Jiasen Lu, Dhruv Batra, Devi Parikh, and Stefan Lee.
\newblock 2019.
\newblock Vilbert: Pretraining task-agnostic visiolinguistic representations
  for vision-and-language tasks.
\newblock In {\em Advances in Neural Information Processing Systems}, pages
  13--23.

\bibitem[\protect\citename{Mehta and Zhu}2009]{Mehta2009}
Ravi Mehta and Rui Zhu.
\newblock 2009.
\newblock Blue or red? exploring the effect of color on cognitive task
  performances.
\newblock {\em Science}, 323:1226–1229.

\bibitem[\protect\citename{Mikolov \bgroup et al.\egroup
  }2013]{mikolov2013distributed}
Tomas Mikolov, Ilya Sutskever, Kai Chen, Greg~S Corrado, and Jeff Dean.
\newblock 2013.
\newblock Distributed representations of words and phrases and their
  compositionality.
\newblock In {\em Advances in neural information processing systems}, pages
  3111--3119.

\bibitem[\protect\citename{Miller}1998]{miller1998wordnet}
George~A Miller.
\newblock 1998.
\newblock {\em WordNet: An electronic lexical database}.
\newblock MIT press.

\bibitem[\protect\citename{P{\'e}rez and Granger}2007]{perez2007iPython}
Fernando P{\'e}rez and Brian~E Granger.
\newblock 2007.
\newblock Ipython: a system for interactive scientific computing.
\newblock {\em Computing in science \& engineering}, 9(3):21--29.

\bibitem[\protect\citename{Riley}1995]{riley1995}
Charles Riley.
\newblock 1995.
\newblock {\em Color Codes: Modern Theories of Color in Philosophy, Painting
  and Architecture, Literature, Music, and Psychology}.
\newblock UPNE.

\bibitem[\protect\citename{Safdar \bgroup et al.\egroup
  }2017]{safdar2017perceptually}
Muhammad Safdar, Guihua Cui, Youn~Jin Kim, and Ming~Ronnier Luo.
\newblock 2017.
\newblock Perceptually uniform color space for image signals including high
  dynamic range and wide gamut.
\newblock {\em Optics express}, 25(13):15131--15151.

\bibitem[\protect\citename{Socher \bgroup et al.\egroup
  }2014]{socher2014grounded}
Richard Socher, Andrej Karpathy, Quoc~V Le, Christopher~D Manning, and Andrew~Y
  Ng.
\newblock 2014.
\newblock Grounded compositional semantics for finding and describing images
  with sentences.
\newblock {\em Transactions of the Association for Computational Linguistics},
  2:207--218.

\bibitem[\protect\citename{Tsvetkov \bgroup et al.\egroup
  }2014]{tsvetkov-etal-2014-metaphor}
Yulia Tsvetkov, Leonid Boytsov, Anatole Gershman, Eric Nyberg, and Chris Dyer.
\newblock 2014.
\newblock Metaphor detection with cross-lingual model transfer.
\newblock In {\em Proceedings of the 52nd Annual Meeting of the Association for
  Computational Linguistics (Volume 1: Long Papers)}, pages 248--258,
  Baltimore, Maryland, June. Association for Computational Linguistics.

\bibitem[\protect\citename{Virtanen \bgroup et al.\egroup
  }2020]{virtanen2020scipy}
Pauli Virtanen, Ralf Gommers, Travis~E Oliphant, Matt Haberland, Tyler Reddy,
  David Cournapeau, Evgeni Burovski, Pearu Peterson, Warren Weckesser, Jonathan
  Bright, et~al.
\newblock 2020.
\newblock Scipy 1.0: fundamental algorithms for scientific computing in python.
\newblock {\em Nature methods}, 17(3):261--272.

\bibitem[\protect\citename{Walt \bgroup et al.\egroup }2011]{walt2011numpy}
St{\'e}fan van~der Walt, S~Chris Colbert, and Gael Varoquaux.
\newblock 2011.
\newblock The numpy array: a structure for efficient numerical computation.
\newblock {\em Computing in science \& engineering}, 13(2):22--30.

\bibitem[\protect\citename{Winter}2019]{Winter2019}
Bodo Winter.
\newblock 2019.
\newblock {\em Sensory Linguistics}.
\newblock John Benjamins Publishing Company.

\bibitem[\protect\citename{{Şenel} \bgroup et al.\egroup }2018]{senel_interp}
L.~K. {Şenel}, İ. {Utlu}, V.~{Yücesoy}, A.~{Koç}, and T.~{Çukur}.
\newblock 2018.
\newblock Semantic structure and interpretability of word embeddings.
\newblock {\em IEEE/ACM Transactions on Audio, Speech, and Language
  Processing}, 26(10):1769--1779.

\end{thebibliography}
\bibliographystyle{coling2020}

\clearpage

\appendix

\section{\texttt{comp-syn} Code}

\texttt{comp-syn} is an open source Python package, available on GitHub and downloadable through PyPi at \MYhref{https://github.com/comp-syn/comp-syn}{https://github.com/comp-syn/comp-syn}. It uses \texttt{NumPy} \cite{walt2011numpy} and \texttt{scipy} \cite{jones2001scipy} for computational purposes, \texttt{matplotlib} \cite{hunter2007matplotlib} for visualisations, and \texttt{Pillow} \cite{clark2018pillow} to load and manipulate images in Python. We provide documentation 
and extensive examples of code use through \texttt{Jupyter/IPython} notebooks \cite{perez2007iPython}.

With a focus on community-based development, our code is PEP8 compliant, heavily commented and documented, includes continuous code integration and testing via Travis CI, and uses GitHub as a platform for issue tracking and ideation. \texttt{comp-syn} is structured so that it is easily compatible with contemporary deep learning, natural language processing and computational linguistics Python packages, and particularly with \texttt{NumPy} and \texttt{SciPy} as standard tools for scientific computing \cite{virtanen2020scipy}. 

\texttt{comp-syn} includes the following modules:

\begin{itemize}
    \item \texttt{datahelper.py} - loader functions for scraping and organising image data;
    \item \texttt{analysis.py} - analysis functions for computing word--color embeddings;
    \item \texttt{logger.py} - logger functions;
    \item \texttt{visualisation.py} - visualisation functions;
    \item \texttt{vectors.py} - word--color embedding loader and manager;
    \item \texttt{wordnet-functions.py} - \texttt{WordNet} functions to facilitate web scraping and linking with concreteness measures.
\end{itemize}

\section{Data Distribution}

Along with the software, we also release word--color embeddings for the nearly 40,000 English language words. Our corpus of word--color embeddings is linked to crowd-sourced human judgments of concept concreteness from Brysbaert et al.\ \shortcite{brysbaert2014concreteness}, and terms used for analysis are also linked to \texttt{WordNet} \cite{miller1998wordnet}. These embeddings can easily be downloaded and loaded into Python pipelines for various natural language processing tasks, including jointly training or enhancing distributional semantics models. The data is stored in the \texttt{json} file \texttt{concreteness-color-embeddings.json} and can be loaded using the \texttt{vectors.py} module class \texttt{LoadVectorsFromDisk}. Once loaded, each \texttt{Vector} object has the following information based on the top $100$ Google Image results for each word:

\begin{itemize}
    \item \texttt{rgb-dist} - the average \texttt{RGB} distribution, computed in $8$ evenly-segmented bins;
    \item \texttt{jzazbz-dist} - the average perceptually uniform $J_z A_z B_z$ distribution, computed in $8$ evenly-segmented bins;
    \item \texttt{jzazbz-dist-std} - the standard deviation over the perceptually uniform $J_z A_z B_z$ distribution, computed in $8$ evenly-segmented bins;
    \item \texttt{colorgram} - a composite image created by pixel-wise averaging over the top $100$ images, representing the word's ``average image'';
    \item \texttt{rgb-vector} - the mean \texttt{RGB} coordinates, averaged over the top $100$ images;
    \item \texttt{jzazbz-vector} - the mean $J_z A_z B_z$ coordinates, averged over the top $100$ images
    \item \texttt{concreteness-mean} - the mean of the concreteness scores for the word from \cite{brysbaert2014concreteness};
    \item \texttt{concreteness-sd} - the standard deviation of the concreteness scores for the word from \cite{brysbaert2014concreteness}.
\end{itemize}

\section{Methods}
\label{methods}

\subsection{Google Image Search}

To generate word--color embeddings, we collect the top $100$ Google Image search results for each of the 40,000 terms in our analysis. The Google Image searches were run from 10 servers running in a commercial datacenter in New York, USA. These servers were created and used only for this experiment, so that results would not be overly-personalized. Additional search parameters were included as query strings \texttt{safe=off\&site=\&tbm=isch\&source=hp\&gs\_l=img}.

\subsection{Word--Color Embeddings}

We use the \texttt{PIL} Python module to convert each image into an $m\times n\times 3$ array of sRGB values, where $m$ and $n$ are the intrinsic image dimensions. For computational efficiency, we then compress each image into an anti-aliased $300\times 300\times 3$ array. Next, we transform sRGB pixel values into their counterparts in the perceptually uniform $J_z A_z B_z$ colorspace.\footnote{\texttt{comp-syn} efficiently computes color embeddings in multiple colorspaces if desired, including more traditional (non-perceptually uniform) options such as \texttt{RGB} and \texttt{HSV}.} Unlike in standard colorspace, Euclidean distances in $J_z A_z B_z$ coordinates linearly correspond to differences in human color perception \cite{safdar2017perceptually}. Moreover, our use of the $J_zA_zB_z$ colorspace rather than a standard colorspace like RGB increases the semantic coherence of our word--color embeddings \cite{guilbeault2020color}.

We measure the color distribution of each image in 8 evenly-segmented $J_z A_z B_z$ subvolumes spanning the range of $J_z A_z B_z$ coordinates that maps to all possible RGB tuples: $J_z\in [0,0.167]$, $A_z\in [-0.1,0.11]$,~$B_z\in [-0.156,0.115]$. Next, we average $J_z A_z B_z$ distributions over all $100$ images for each term to obtain an aggregate, 8-dimensional color mean embedding. We also compute the standard deviation over the $100$ images in each $J_zA_zB_z$ subvolumes to obtain an aggregate, 8-dimensional color variability embedding. These color mean and variability embeddings can be concatenated to create a 16-dimensional embedding. The details of our compression, binning, and averaging steps do not affect our results \cite{guilbeault2020color}.

To compare word--color embeddings, we use the Jensen-Shannon (JS) divergence to measure the similarity of aggregate $J_zA_zB_z$ distributions. In particular, for $J_zA_zB_z$ distributions $C_i$ and~$C_j$ associated with terms $i$ and $j$, the JS divergence is given by
\begin{equation}
    D_{\mathrm{JS}}(C_1~||~C_2) \equiv \frac{1}{2}\left[D_{\mathrm{KL}}(C_1~||~\bar{C}_{12}) + D_{\mathrm{KL}}(C_2~||~\bar{C}_{12})\right],\label{eq:js}
\end{equation}
where $D_{\rm{KL}}$ is the Kullback-Leibler divergence and $\bar{C}_{12}\equiv (C_1+C_2)/2$. The JS divergence is a measure of the distance between two color distributions, such that lower values correspond to more similar distributions in perceptually uniform colorspace. We choose this metric because it is a well-defined distance measure that satisfies the triangle inequality and allows us to avoid undefined values associated with empty $J_zA_zB_z$ bins. Terms with relatively high mutual JS divergences usually exhibit \emph{colorgrams} with perceptibly different average colors.

\section{Best Practices for Usage}

We caution that our word--color embeddings are \emph{distributions} rather than \emph{vectors}. Thus, their components are positive semidefinite, and different embeddings must be compared using similarity measures designed for distributions such as JS divergence. On the other hand, \texttt{word2vec} embeddings are vectors with components that can be positive or negative, and are compared using similarity measures designed for vectors such as cosine similarity. Importantly, unlike \texttt{word2vec} embeddings, our word--color embeddings cannot be composed by vector addition. We are exploring algebraic techniques for word composition in colorspace; these techniques must respect the underlying mathematical structure of $J_zA_zB_z$ (and other) colorspaces, which are not closed under standard binary operations like addition or multiplication.

\end{document}